\documentclass[sigconf]{acmart}
\renewcommand\footnotetextcopyrightpermission[1]{} 
\AtBeginDocument{%
  \providecommand\BibTeX{{%
    \normalfont B\kern-0.5em{\scshape i\kern-0.25em b}\kern-0.8em\TeX}}}

\setcopyright{none}
\copyrightyear{2020}
\acmYear{2020}
\acmDOI{10.1145/1122445.1122456}

\acmConference[KDD '20]{KDD '20: ACM SIGKDD Conference}{August 22--27, 2020}{San Diego, United States}
\acmBooktitle{KDD '20: ACM SIGKDD Conference,
August 22--27, 2020, San Diego, United States}
\acmPrice{15.00}
\acmISBN{978-1-4503-9999-9/18/06}

\usepackage{epsfig,amsmath,amsfonts,epsfig,multirow,makecell,caption,soul,csquotes,color,wrapfig,subcaption,mathtools,bm,spverbatim,booktabs,tcolorbox,diagbox,todonotes}
\usepackage[e]{esvect}

\usepackage{caption}
\makeatletter
\newif\if@restonecol
\makeatother

\usepackage[boxed, ruled, vlined, linesnumbered]{algorithm2e}
\SetKwRepeat{Do}{do}{while}

\newenvironment{changemargin}[2]{\begin{list}{}{
	\setlength{\topsep}{0pt}\setlength{\leftmargin}{0pt}
	\setlength{\rightmargin}{0pt}
	\setlength{\listparindent}{\parindent}
	\setlength{\itemindent}{\parindent}
	\setlength{\parsep}{0pt plus 1pt}
	\addtolength{\leftmargin}{#1}\addtolength{\rightmargin}{#2}
	}\item}
	{\end{list}}

\newenvironment{mitemize}{
	\begin{changemargin}{-3pt}{-0cm}
	\vspace{-10pt}
	\hspace{-5pt}
	\begin{itemize}
	\setlength{\itemsep}{3pt}}
	{\end{itemize}
	\vspace{2pt}
	\end{changemargin}}

\newcommand{\ssup}[2]{{#1}^{\scaleobj{0.8}{#2}}}
\newcommand{\ssub}[2]{{#1}_{\scaleobj{0.8}{#2}}}
\newcommand{\sboth}[3]{{#1}_{\scaleobj{0.8}{#2}}^{\scaleobj{0.8}{#3}}}

\usepackage[first=0,last=9]{lcg}
\usepackage{colortbl}
\definecolor{Gray}{gray}{0.8}

\usepackage{hyperref}

\newcommand{\msec}[1]{\S\,\ref{#1}}
\newcommand{\mref}[1]{\,\ref{#1}}
\newcommand{\meq}[1]{Eqn\,(\ref{#1})}
\newcommand{\mcite}[1]{\,\cite{#1}}
\newcommand{\met}{\;\textit{et al.}\xspace}

\usepackage{scalerel}[2016/12/29]

\makeatletter
\providecommand{\leadsfrom}{%
  \mathrel{\mathpalette\reflect@squig\relax}%
}
\newcommand{\reflect@squig}[2]{%
  \reflectbox{$\m@th#1\leadsto$}%
}
\makeatother

\newcommand{\bx}{\ssub{x}{\circ}}

\newcommand{\ax}{\ssub{x}{*}}
\newcommand{\ay}{\ssub{c}{*}}

\newcommand{\dnn}{DNN\xspace}
\newcommand{\dnns}{DNNs\xspace}

\def\eqref#1{equation~\ref{#1}}

\def\1{\bm{1}}

\DeclareMathAlphabet{\mathsfit}{\encodingdefault}{\sfdefault}{m}{sl}
\SetMathAlphabet{\mathsfit}{bold}{\encodingdefault}{\sfdefault}{bx}{n}

\def\gC{{\mathcal{C}}}

\def\gX{{\mathcal{X}}}

\def\sE{{\mathbb{E}}}

\def\sR{{\mathbb{R}}}

\DeclareMathOperator{\sign}{sgn}

\usepackage{amssymb}

\newtheorem{prop}{Proposition}[section]

\newcommand{\system}{\texttt{AdvMind}\xspace}

\newcommand{\amr}{$\ssup{\texttt{AdvMind}}{\texttt{R}}$\xspace}

\newcommand{\amrp}{$\ssup{\texttt{AdvMind}}{\texttt{RP}}$\xspace}

\begin{document}

\title{\system: Inferring Adversary Intent of Black-Box Attacks}

\author{Ren Pang}
\email{rbp5354@psu.edu}
\affiliation{%
  \institution{Pennsylvania State University}
}

\author{Xinyang Zhang}
\email{xqz5366@psu.edu}
\affiliation{%
  \institution{Pennsylvania State University}
}

\author{Shouling Ji}
\email{sji@zju.edu.cn}
\affiliation{%
  \institution{Zhejiang University, Ant Financial}
}

\author{Xiapu Luo}
\email{csxluo@comp.polyu.edu.hk}
\affiliation{%
  \institution{Hong Kong Polytechnic University}
}

\author{Ting Wang}
\email{inbox.ting@gmail.com}
\affiliation{%
  \institution{Pennsylvania State University}
}



\begin{abstract}
Deep neural networks (DNNs) are inherently susceptible to adversarial attacks even under black-box settings, in which the adversary only has query access to the target models. In practice, while it may be possible to effectively detect such attacks (e.g., observing massive similar but non-identical queries), it is often challenging to exactly infer the adversary intent (e.g., the target class of the adversarial example the adversary attempts to craft) especially during early stages of the attacks, which is crucial for performing effective deterrence and remediation of the threats in many scenarios.

In this paper, we present \system, a new class of estimation models that infer the adversary intent of black-box adversarial attacks in a {\em robust} and {\em prompt} manner. Specifically, to achieve robust detection, \system accounts for the adversary adaptiveness such that her attempt to conceal the target will significantly increase the attack cost (e.g., in terms of the number of queries); to achieve prompt detection, \system proactively synthesizes plausible query results to solicit subsequent queries from the adversary that maximally expose her intent. Through extensive empirical evaluation on benchmark datasets and state-of-the-art black-box attacks, we demonstrate that on average \system detects the adversary intent with over 75\% accuracy after observing less than 3 query batches and meanwhile increases the cost of adaptive attacks by over 60\%. We further discuss the possible synergy between \system and other defense methods against black-box adversarial attacks, pointing to several promising research directions.
\end{abstract}


\maketitle

\section{Introduction}

The recent advances in deep learning\mcite{nat:dl} have led to breakthroughs in a range of long-standing machine learning tasks (e.g., image classification\mcite{Deng:2009:cvpr}, natural language processing\mcite{rajpurkar:squad}, and even playing Go\mcite{silver:nature:alphago}), enabling many scenarios previously considered strictly experimental. However, it is well known that deep neural network (\dnn) models are inherently vulnerable to adversarial inputs -- those maliciously crafted samples to force the target {\dnns} to misbehave\mcite{szegedy:iclr:2014} -- which significantly hinder their use in security-sensitive domains. Typically, adversarial inputs are crafted by carefully perturbing legitimate samples under the guidance of the gradient information of target DNNs (i.e., ``white-box'' attacks)\mcite{goodfellow:fsgm,carlini:sp:2017}.

Meanwhile, many cloud-based service providers, including Amazon, Google, Microsoft, BigML, and others all have arisen to provide Machine Learning-as-a-service (MLaaS) platforms. On such platforms, increasingly many commercial and proprietary DNN models are being deployed with publicly accessible interfaces (``predictive APIs''), which allow users to query the backend models with inputs of interests and charge users on a pay-per-query basis. For instance, the \textsc{Clarifai} NSFW\footnote{\url{https://www.clarifai.com/models}} (``not safe for work'') detection API returns probability scores that a given image contains nudity. The first 2,500 queries are free, it then charges \$2.4 per 1,000 queries.

\begin{figure}
    \centering
    \epsfig{file=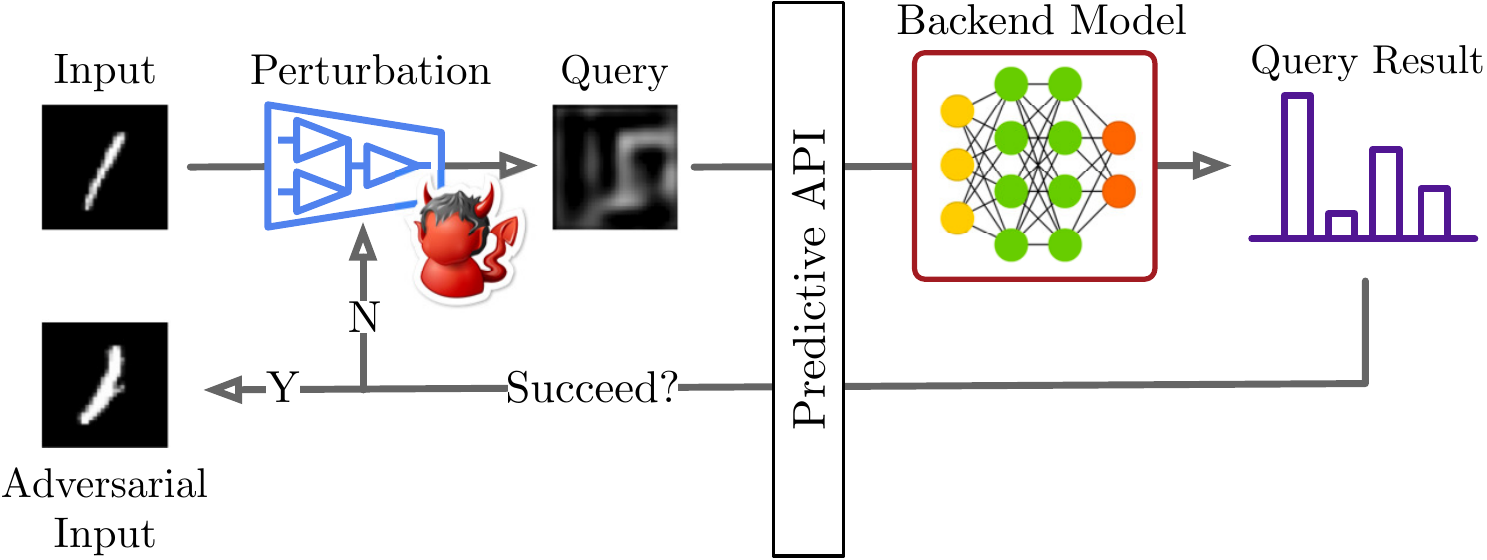, width=75mm}
    \caption{Illustration of black-box adversarial attacks. \label{fig:blackbox}}
\end{figure}

Under such settings, the adversary must (i) construct adversarial inputs with only query access to the target DNNs (i.e., without gradient information) and (ii) also minimize the attack cost in terms of the number of queries, which spurs the research on ``black-box'' adversarial attacks. The existing black-box attacks in the literature can be categorized in two  classes. The first one trains a surrogate \dnn to emulate the target model and attacks the surrogate model using first-order white-box methods\mcite{papernot2017practical}. However, it often requires on the order of millions of queries to construct surrogate models for DNNs used in practice\mcite{Ilyas:2018:icml}; the second one  constructs adversarial inputs by estimating the gradient of target \dnns via coordinate-wise finite difference methods\mcite{chen2017zoo,Narodytska:2017:cvpr,Ilyas:2018:icml}, as illustrated in Figure\mref{fig:blackbox}. Due to its practical feasibility, in the following, we focus on the second class of black-box adversarial attacks.





Given their reliance on zeroth order methods, such black-box attacks often require issuing a number of queries in the vicinity of a point of interest to accurately estimate its gradient information. Thus, it may be straightforward to detect the undergoing black-box attacks (e.g., by observing massive similar but non-identical queries\mcite{blackbox-detection}). Yet, it is often challenging to accurately infer the adversary's intent -- the target class of the adversarial input the adversary attempts to craft -- especially during early stages of the attack. Identifying the adversary's target in many scenarios,
such as the person the adversary impersonates in face recognition\mcite{Biggio:2012:spr}, the insurance category the adversary claims in healthcare fraud\mcite{medical-insurance}, and the legitimate source the adversary disguises as in fake news detection\mcite{Liang:2016:www}, is critical for deploying proper mitigation strategies or performing prompt remediation against such threats.

Yet, despite the plethora of prior work on black-box adversarial attacks, the research on understanding the adversary's intent is still fairly limited. To the best of our knowledge, this work represents a solid step towards bridging this gap. We present \system, a novel framework for early-stage detection of the adversary's intent in black-box adversarial attacks. At a high level, \system achieves {\em robust} and {\em prompt} inference of adversary intent with two key strategies:


\vspace{2pt}
{\em Robust intent estimation --} Taking into account that the adversary may purposely issue fake queries to conceal her true intent, \system employs an mean estimation model that is (i) reliable even in the presence of adversarial noise and (ii) agnostic to the underlying distribution from which the adversary samples the queries. Furthermore, by leveraging the fact that camouflage requires significantly increasing the attack cost (e.g., the number of queries issued), \system creates for the adversary the dilemma between intent disclosure and attack cost.

\vspace{2pt}
{\em Proactive intent solicitation --} To achieve early-stage detection, \system proactively synthesizes plausible query results to solicit subsequent queries from the adversary that maximally exposes her intent. Moreover, generated by slight perturbation on first-order information, the synthetic query results retain sufficient accuracy for legitimate use but can cause significant deviation if the results are used for gradient estimation as in black-box adversarial attacks, which further deters the adversary from successful attacks.


\vspace{2pt}
We extensively evaluate the efficacy of \system with respect to benchmark datasets, popular DNNs, and state-of-the-art black-box attacks. We show that across all cases, \system manages to detect the adversary intent with over 75\% accuracy after observing less than 3 query batches and meanwhile increases the query cost of adaptive attacks by over 60\%. We further discuss the synergy between \system and existing defenses against black-box adversarial attacks, pointing to several promising research directions.\footnote{The code and data used in the paper is released at \url{https://github.com/alps-lab/advmind}.}




\section{Fundamentals}
\label{sec:background}

We begin with introducing a set of concepts and assumptions used throughout the paper. The important symbols and notations are summarized in Table\mref{tab:symbol} in Appendix.

\subsection{Threat Models}

In the following, we primarily focus on predictive tasks (e.g., image classification\mcite{Deng:2009:cvpr}). Under this setting, a DNN $f$ represents a function $f: \gX \rightarrow \gC$, which assigns a given input $x \in \gX$ to one of a set of predefined classes $\gC$, $f(x) = c \in \gC$. \dnns are inherently vulnerable to adversarial inputs, which are maliciously crafted to force target \dnns to misbehave\mcite{Dalvi:2004:kdd,szegedy:iclr:2014,moosavi:cvpr:2017,madry:iclr:2018}. Here we focus on the setting of targeted attacks. Specifically, an adversarial input $x_*$ is often generated by slightly modifying a benign input $\bx$, with the objective of forcing $f$ to misclassify $\ax$ into a target class $\ay$, $f(\ax) = \ay \neq f(\bx)$. To ensure that $\ax$ is perceptually similar to $\bx$, the perturbation is constrained to a set of allowed perturbations (e.g., a norm ball $\mathcal{B}_\varepsilon(\bx) = \{x | \|x - \bx \|_\infty \leq \varepsilon\}$).

Consider project gradient descent (PGD)\mcite{madry:iclr:2018}, a universal first-order adversarial attack, as an example. At a high level, it is implemented as a sequence of project gradient descent steps on the negative loss function:
\hspace{-15pt}
\begin{align}
  \label{eq:pgd}
  \ssup{x}{(i+1)} = \Pi_{\mathcal{B}_\varepsilon(\bx)} \left(\ssup{x}{(i)} - \alpha \sign\left(\nabla\ell\left(f(\ssup{x}{(i)}), \ay\right)\right)\right)
\end{align}
where $\Pi$ denotes the projection operator, $\alpha$ ($\alpha \geq 0$) represents the learning rate, the loss function $\ell$ measures the difference of the model prediction $f(x)$ and the class $\ay$ desired by the adversary (e.g., cross entropy), and $\ssup{x}{(0)}$ is initialized as $\bx$.


%
%
%
%
%
%
%
%
%
%
%
%

\begin{algorithm}{\small
  \KwIn{original input $\ssup{x}{(0)}=\bx$, target class $\ay$, target DNN $f$, learning rate $\alpha$, norm constraint $\varepsilon$, and number of iterations $n_\mathrm{iter}$}
  \KwOut{adversarial input $\ax$}
  \For{$i = 1, \ldots, n_\mathrm{iter}$}
  {
    $\hat{g}(\ssup{x}{(i)})\leftarrow \text{GradEstimate}(\ssup{x}{(i)}, \ay)$\;
    $ \ssup{x}{(i+1)} \leftarrow \Pi_{\mathcal{B}_\varepsilon(\bx)}\left(\ssup{x}{(i)} -\alpha \sign\left(\hat{g}(\ssup{x}{(i)})\right)\right)$\;
    \lIf{$f(\ssup{x}{(i+1)}) = \ay$}{\Return $\ssup{x}{(i+1)}$ as $\ax$}
  }
  \Return Fail\;
  \caption{\textsf{Black-Box Attack} \label{alg:blackbox}}}
\end{algorithm}

Under the black-box setting, the adversary has only query access to the target DNN $f$ (i.e., without the gradient information), as sketched in Algorithm\mref{alg:blackbox}. Note that here we focus on an account-oriented setting, where users must create an account before they can query the model\mcite{blackbox-detection}. The adversary typically uses coordinate-wise finite difference methods to estimate the gradient and then applies gradient descent as in \meq{eq:pgd} to generate adversarial inputs $\ax$. Different black-box adversarial attacks mainly differ in their gradient estimation method {\sf GradEstimate}$(x, \ay)$ in Algorithm\mref{alg:blackbox}. In this work, we consider three representative black-box attacks.

To clarify, $f(x)$ returns the classification label $c$ of image $x$ in the above content. However, we'll introduce different attack methods in the following section, within which $f(x)$ is the probability confidence vector after softmax layer. For simplicity, in the remainder of the paper, $f(x)$ stands for the classification label in $\ell(f(x),\ay)$, and stands for the probability vector when it's alone.

\subsubsection*{\bf Zeroth Order Natural Evolutionary Strategy}
In\mcite{Ilyas:2018:icml}, Ilyas\met apply natural evolutionary strategies (NES), a derivative-free optimization method\mcite{Wierstra:2014:jml} to estimate the gradient of $\hat{g}(\ssup{x}{(i)})$. Specifically, NES generates $n_\mathrm{query}$ data points in the neighborhood of $\ssup{x}{(i)}$ by sampling from a normal distribution, retrieves their predictions from the target DNN $f$, and estimates the gradient $\hat{g}(\ssup{x}{(i)})$ as follow:

\begin{align}
  \label{eq:nes1}
  \hat{g}(\ssup{x}{(i)}) = \frac{1}{\sigma n_\mathrm{query}} \sum_{j=1}^{n_\mathrm{query}} f\left(\ssup{x}{(i)}+ \sigma \ssup{u}{(j)}\right)\ssup{u}{(j)}
\end{align}
where each sample $\ssup{u}{(j)}$  is sampled from the standard normal distribution $\mathcal{N}(0, I)$, and $\sigma$ is the sampling variance.

To accelerate the sampling, it is possible to generate only half of the samples from the distribution and set the other half symmetrically. The gradient estimate is then formulated as follows:
\begin{align}
  \hspace{-10pt}
  \label{eq:nes2}
  \hat{g}(\ssup{x}{(i)}) = \frac{1}{\sigma n_\mathrm{query}} \sum_{j=1}^{[n_\mathrm{query}/2]} \left(f\left(\ssup{x}{(i)}+\sigma \ssup{u}{(j)}\right)-f\left(\ssup{x}{(i)}-\sigma \ssup{u}{(j)}\right)\right)\ssup{u}{(j)}
\end{align}

\vspace{2pt}
\subsubsection*{\bf Zeroth Order signSGD}

In\mcite{liu2018signsgd}, Liu\met integrate signSGD\mcite{pmlr-v80-bernstein18a}, a compressed optimization method that only uses the sign of gradient, with zeroth order optimization\mcite{chen2017zoo}, and propose one-sided and two-sided
estimators of $\hat{g}(\ssup{x}{(i)})$. Specifically, the one-sided estimator is given by:
\begin{align}
  \label{eq:sgd1}
  \begin{split}
    \hat{g}(\ssup{x}{(i)})  = & \frac{1}{\sigma n_\mathrm{query}} \sum_{j=1}^{n_\mathrm{query}} \left(f\left(\ssup{x}{(i)}+\sigma \ssup{u}{(j)}\right)-f\left(\ssup{x}{(i)}\right)\right)\ssup{u}{(j)} \\
    =& \frac{1}{\sigma n_\mathrm{query}} \sum_{j=1}^{n_\mathrm{query}} f\left(\ssup{x}{(i)}+ \sigma \ssup{u}{(j)}\right)\ssup{u}{(j)} - \frac{f\left(\ssup{x}{(i)}\right)}{\sigma n_\mathrm{query}} \sum_{j=1}^{n_\mathrm{query}} \ssup{u}{(j)}
  \end{split}
\end{align}
where the NES estimator in \meq{eq:nes1} can be used to compute the first term above. Also note that since the expectation of each sample $\ssup{u}{(j)}$ is 0, \meq{eq:sgd1} and \meq{eq:nes1} share the same expectation. Yet, compared with the NES estimator, the second term in \meq{eq:sgd1} reduces the error caused by the variance of $\ssup{u}{(j)}$, leading to faster convergence with the number of samples.

The two-sided estimator in\mcite{liu2018signsgd} is essentially identical to the improved NES estimator in \meq{eq:nes2}.

\subsubsection*{\bf Zeroth Order Hessian-Aware}

Note that the NES and signSGD estimators only use first-order information in estimating $\hat{g}(\ssup{x}{(i)})$. In\mcite{HessAware}, Haishan \met argue that it is more efficient to estimate $\hat{g}(\ssup{x}{(i)})$ if the second-order information (Hessian) is considered. They propose the following gradient estimator:
\begin{align}
  \hspace{-10pt}
  \hat{g}(\ssup{x}{(i)}) =
  \frac{1}{\sigma n_\mathrm{query}} \sum_{j=1}^{ n_\mathrm{query}} \left(f\left(\ssup{x}{(i)}+ \sigma \Tilde{H}^{-\frac{1}{2}} \ssup{u}{(j)}\right)-f\left(\ssup{x}{(i)}\right)\right)\Tilde{H}^{-\frac{1}{2}}\ssup{u}{(j)}
\end{align}
where $\Tilde{H}^{-\frac{1}{2}}$ is the Cholesky inverse of the Hessian estimate $\Tilde{H}$. Further, $\Tilde{H}$ is estimated as follows:
\begin{align}
  \Tilde{H}= & \frac{\sum^{ n_\mathrm{query}}_{j=1}\left| f\left(\ssup{x}{(i)}+ \sigma \ssup{u}{(j)}\right)+f\left(\ssup{x}{(i)} - \sigma \ssup{u}{(j)}\right)-2f\left(\ssup{x}{(i)}\right)\right| \ssup{u}{(j)} \otimes\ssup{u}{(j)}}{2 \sigma^2 n_\mathrm{query}}
\end{align}
where $\otimes$ represents the outer product of two vectors. Note that to ensure the feasibility of Cholesky decomposition and invertibility, a small identity matrix (e.g., $\tau I$) is often added to $\tilde{H}$.

\subsection{Overview of \system}

Despite their apparent variations, the query-based black-box adversarial attacks share common patterns. Each attack consists of a sequence of iterations. At the $i$-th iteration, the adversary issues queries regarding the current input $\ssup{x}{(i)}$, which we refer to as the query of interest (QOI), and the auxiliary data points in the vicinity of $\ssup{x}{(i)}$, which are used to estimate $\hat{g}(\ssup{x}{(i)})$.

\begin{figure}[!ht]
  \centering
  \epsfig{file = 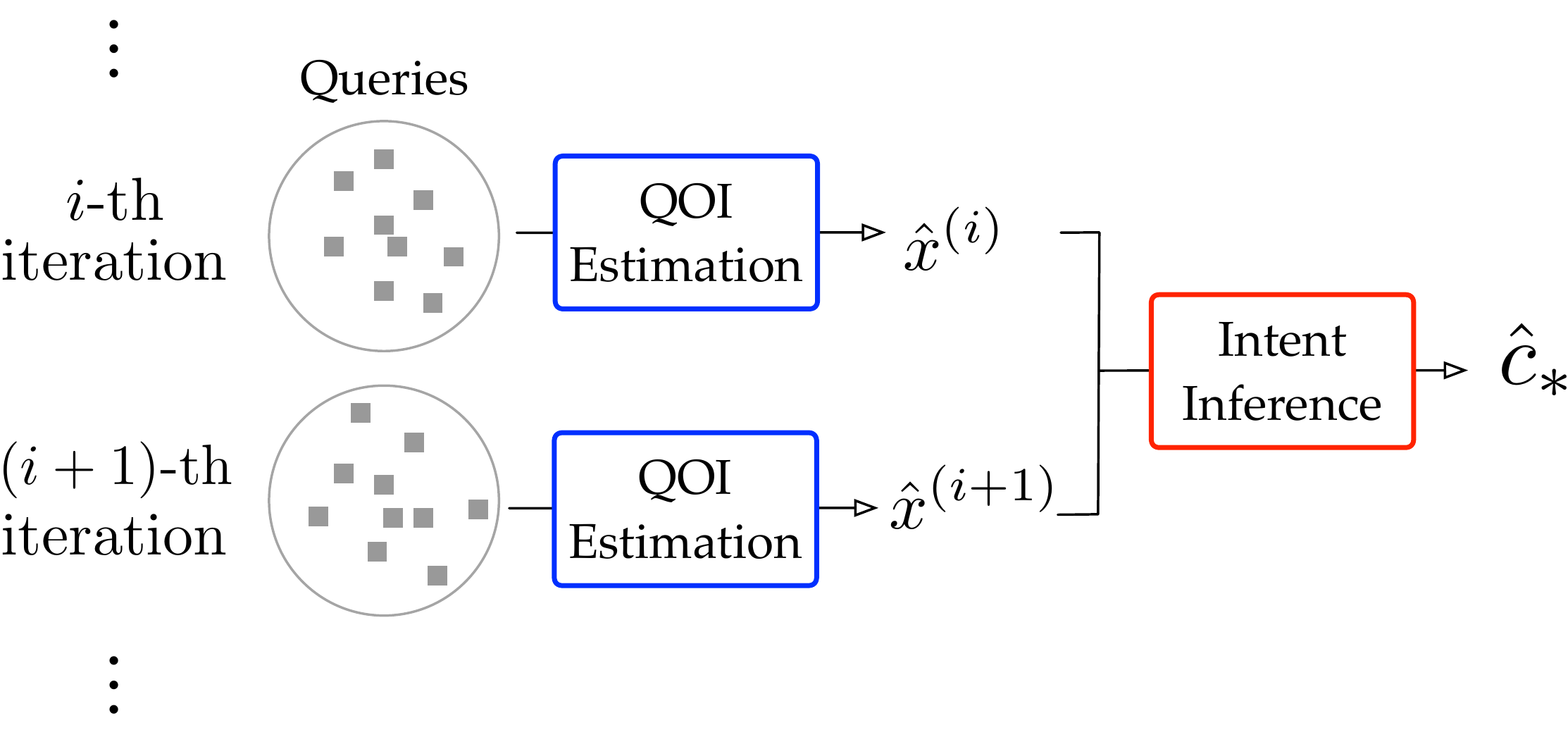, width = 80mm}
  \caption{Illustration of \system framework. \label{fig:framework}}
\end{figure}

Therefore, to infer the adversary's intent (i.e., the target class $\ay$), \system adopts the following strategies: 
(i) grouping the incoming queries into a sequence of query batches, each corresponding to an iteration (the feasibility of grouping queries into different iterations is detailed in Appendix B); (ii) estimating the QOI at each iteration -- As the auxiliary queries appear in the close vicinity of $\ssup{x}{(i)}$, it is feasible to accurately estimate $\ssup{\hat{x}}{(i)}$ by clustering the queries; and (iii) estimating the gradient direction followed by the adversary -- By linking consecutive QOIs (e.g., $\ssup{\hat{x}}{(i)}$ and $\ssup{\hat{x}}{(i+1)}$), it is feasible to infer the gradient direction the adversary is following to craft the adversarial input. Intuitively, as the adversary attempts to minimize the loss function with respect to the target class $\ay$, the gradient tends to point to that direction. Further, by combining the inference at multiple iterations, it is possible to reliably infer the adversary's intent $\ay$. This scheme is illustrated in Figure\mref{fig:framework}.

Yet, this na\"{i}ve inference model suffers the following drawbacks.

First, at each iteration, the adversary may attempt to inject fake queries (i.e., irrelevant to the QOI) to conceal her intent. As outliers, the fake queries may significantly deviate \system's estimate $\ssup{\hat{x}}{(i)}$ from the ground-truth QOI $\ssup{x}{(i)}$. To account for the possible adversarial noise, \system adopts a robust QOI estimator. With reasonable fake query proportion $p_\mathrm{fake}$ (e.g., $p_\mathrm{fake} \leq 40\%$), the robust estimator is able to estimate the QOI with bounded bias. Further, it creates for the adversary the dilemma between intent disclosure and attack cost (e.g., in terms of the number of queries).

Second, as the adversary performs project descent based on the estimated gradient (Algorithm\mref{alg:blackbox}), the descent direction she follows may deviate from the true gradient direction, which in turn causes errors in \system's inference of the adversary's target class $\ay$. To enable the early-stage inference and minimize the inference uncertainty, \system adopts proactive intent solicitation. Specifically, by slightly perturbing the query answers, \system solicits subsequent queries from the adversary that maximally expose the target class. Note that such perturbation retains sufficient accuracy in the query answers for legitimate use but causes significant deviation for gradient estimation.

We elaborate on the strategies of robust intent estimation and proactive intent solicitation in \msec{sec:estimate} and \msec{sec:solicit} respectively.

\section{QOI Estimation}
\label{sec:estimate}

In this section, we focus on the estimation of query of interest (QOI) at each iteration. We first introduce the na\"{i}ve QOI estimator, which is however vulnerable to adversarial noise (i.e., fake queries), and then present a robust QOI estimator, which bounds the estimation bias under reasonable noise ratio.

\subsection{Na\"{i}ve Estimator}

Recall that at $i$-th iteration, the adversary samples the set of auxiliary queries $\ssup{\gX}{(i)}$ in the vicinity of the QOI $\ssup{x}{(i)}$ following a certain distribution (e.g., Gaussian or uniform). We may therefore estimate $\ssup{\hat{x}}{(i)}$ using the mean or median of the received queries $\ssup{\gX}{(i)}$.\footnote{When the context is clear, we omit the superscript $(i)$ in the notations.} Note that compared with the mean estimate, the median estimate is more robust to outliers but has a larger bias as it does not utilize the data points on edges.

\begin{figure}[!ht]
  \includegraphics[scale=0.5]{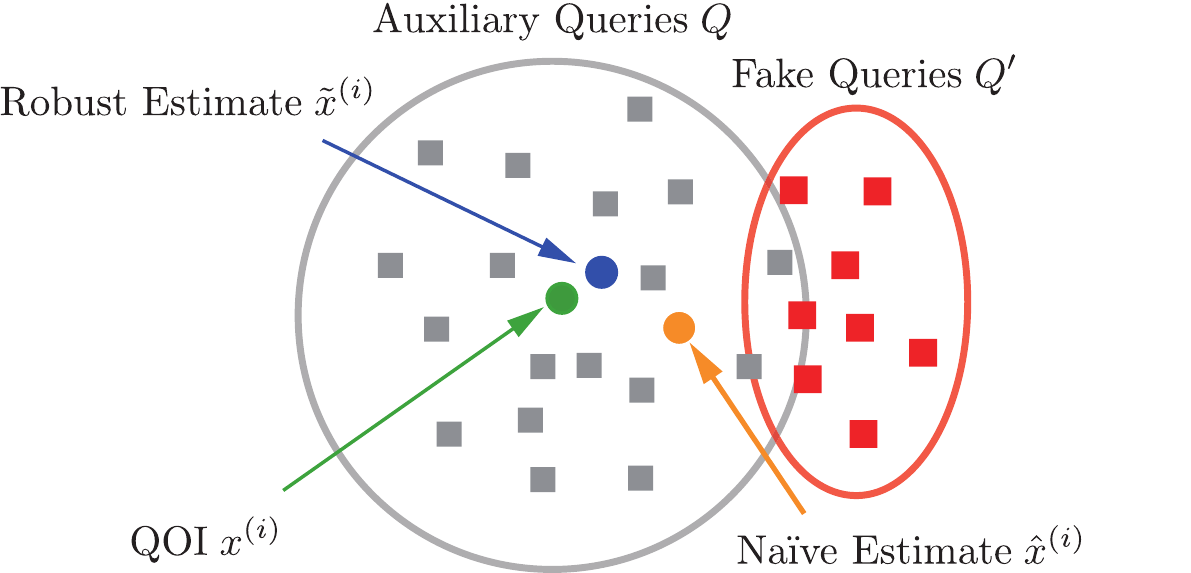}
  \caption{Na\"{i}ve versus robust estimation of QOI in the presence of adversarial noise.}
  \label{fig:robust_figure}
\end{figure}

However, to conceal her intent, the adversary may adaptively inject fake queries into $\gX$ to deviate $\hat{x}$ from the true QOI $x$. As illustrated in Figure\mref{fig:robust_figure}, the presence of such adversarial noise significantly affects the na\"{i}ve estimator, even for the median estimate. The adversary may select from a variety of models to generate fake queries, including random sampling, duplicating existing queries, and query blinding\mcite{blackbox-detection}, which lead to \system's varying estimation error (e.g., measured by $\|\hat{x} - x\|$).

To be concise, in the following we consider Huber's contamination model\mcite{huber1964}, which subsumes a number of models above. Specifically, it assumes that the set of queries $x_1, \dots, x_{n_\mathrm{query}}$ are randomly sampled from a mixture distribution, with the cumulative distribution defined by:
\begin{align}
  \label{huber}
  F=\left(1-p_\mathrm{fake}\right)\mathrm{\Phi}+p_\mathrm{fake} H
\end{align}
where $p_\mathrm{fake}$ is the fraction of fake queries ($0\le p_\mathrm{fake}<0.5$) among all the queries, $H$ is the (unknown) cumulative distribution of fake queries, and $\Phi$ is the cumulative distribution of true auxiliary queries (e.g., Gaussian or uniform).

Next we develop a robust QOI estimator against the influence of fake queries and bounds the estimation bias even under the worst-case distribution $H$.

\subsection{Robust Estimator}
\label{mestimator}

We extend the robust mean estimator\mcite{huber1964,ROUSSEEUW2002741} to the context of QOI estimation. Specifically, given the set of queries $\gX$ ($|\gX| = n_\text{query}$), we estimate the QOI in an iterative manner, in which the $k$-th round estimator $\ssup{\tilde{x}}{(k)}$ is updated as follows:
\begin{align}
  \label{m1}
  \ssup{\tilde{x}}{(k)}=\ssup{\tilde{x}}{(k-1)} +
  \frac{\mathrm{MAD}(\gX)}{\sE_\Phi[\psi^\prime]} \cdot\frac{\sum_{i=1}^{n_\text{query}}\psi\left(\frac{x_i- \ssup{\tilde{x}}{(k-1)}}{\mathrm{MAD}(\gX)}\right)}{n_\text{query}}
\end{align}
where the function $\psi$ is defined as $\psi(x)=(e^x-1)/(e^x+1)$, $\mathrm{MAD}(\gX)$ is the median average deviation of the query set $\gX$. Concretely, with $\Phi$ instantiated as the standard Gaussian distribution, we may estimate the expectation $\sE_\Phi[\psi^\prime]$ as follows:
\begin{align}
  \sE_\Phi [\psi^\prime]=\int_{-\infty}^{+\infty} \psi^\prime(s)\,\mathrm{d}\Phi(s)\approx0.4132
\end{align}

The value of $\ssup{\tilde{x}}{(0)}$ is initialized as the median of the query set $\gX$. As observed in our empirical evaluation (which is also consistent with the results of \mcite{ROUSSEEUW2002741}), one-iteration estimation $\ssup{\tilde{x}}{(1)}$ is typically accurate enough to approximate the QOI $x$. Thus we set the number of iterations $k = 1$ by default.

\vspace{2pt}
Further, we may also estimate the bias of $\tilde{x}$ under the worst-case distribution of adversarial noise, which is needed for inferring the adversary's target in \msec{sec:solicit}. Specifically, the bias of the $k$-th iteration estimator, $B(\ssup{\tilde{x}}{(k)})$, is updated as:

\begin{align}
  \hspace{-20pt}
  \begin{split}
    B(\ssup{\tilde{x}}{(k)})=&B(\ssup{\tilde{x}}{(k-1)})\\
    +&\frac{\left(1-p_\mathrm{fake}\right)\sE_\Phi[\psi(X-B(\ssup{\tilde{x}}{(k-1)}))]+p_\mathrm{fake}\psi(\infty)}{\sE_\Phi[\psi^\prime]}
  \end{split}
\end{align}
with the initial value given by $B(\ssup{\tilde{x}}{(0)})=\mathrm{\Phi}^{-1}\left(\frac{1}{2(1-p_\mathrm{fake})}\right)$, where $\ssup{\Phi}{-1}$ is the inverse function of the Gaussian distribution.
\section{Intent Inference}
\label{sec:solicit}

In this section, we introduce \system's intent inference model, which based on the QOIs estimated at consecutive iterations, identifies the adversary's target class. Moreover, to infer the adversary's intent at an early attack stage and to minimize the inference uncertainty, \system employs a proactive solicitation strategy, which solicits subsequent queries from the adversary that maximally expose her target class.

\subsection{Passive Inference}

We begin with the passive inference model, which identifies the target class $\ay$ directly from the QOI estimation.

Specifically, assuming $\ssup{\tilde{x}}{(i)}$ and $\ssup{\tilde{x}}{(i+1)}$ as the estimated QOIs at two consecutive iterations, we computes the descent direction vector $\ssup{d}{(i)} = \ssup{\tilde{x}}{(i+1)} - \ssup{\tilde{x}}{(i)}$ and the gradient vector $\ssub{g}{c}(\ssup{\tilde{x}}{(i)})$ at $\ssup{\tilde{x}}{(i)}$ with respect to each class $c \in \gC$, and regards the cosine similarity of $\ssup{d}{(i)}$ and $\ssub{g}{c}(\ssup{\tilde{x}}{(i)})$ as the score of $c$:
\begin{align}
  \mathrm{score}(c) =  \frac{  \ssup{d}{(i)} \cdot \ssub{g}{c}(\ssup{\tilde{x}}{(i)}) }{ \|  \ssup{d}{(i)}\| \|  \ssub{g}{c}(\ssup{x}{(i)})\|}
\end{align}

We further normalize the scores of $\gC$ and transforms them into a probability vector using the softmax function. Let $\ssup{p}{(i)}$ denote the probability vector estimated at the $i$-th iteration. We then infer the most likely target class $\ssub{\hat{c}}{*}$ as the class $c$ leading to the overall largest probability:
\begin{align}
  \ssub{\hat{c}}{*} =  \arg\max_{c \in \gC} \Pi_i \sboth{p}{c}{(i)}
\end{align}
Algorithm\mref{alg:nd} sketches the procedure of adversary intent inference for $n_\text{iter}$ iterations.

\begin{algorithm}{\small
    \KwIn{number of iterations $n_\text{iter}$, QOI estimates $(\ssup{\tilde{x}}{(1)}, \ldots, \ssup{\tilde{x}}{(n_\text{iter})})$, confidence threshold $\kappa$}
    \KwOut{target class estimate $\ssub{\hat{c}}{*}$}
    \tcp{initialization}
    $p \leftarrow \1$\;
    \For{$i = 2, \ldots, n_\text{iter}$}
    {
      $d \leftarrow \ssup{\tilde{x}}{(i+1)} - \ssup{\tilde{x}}{(i)}$\;
      \For{$c\in \gC$}
      {
        \tcp{descent direction estimation}
        $\ssub{g}{c} = \nabla \ell(f\left(\ssup{\tilde{x}}{(i)}\right), c)$\;
        $\ssub{s}{c} \leftarrow  \frac{  d \cdot \ssub{g}{c}}{ \|  d\| \|  \ssub{g}{c}\|}$\;
      }
      \tcp{normalization}
      $p \leftarrow \mathsf(p \odot \mathsf{softmax}(s))$\;
      \lIf{$\max_c\ssub{p}{c} \geq \kappa$}
      {
        \Return $\arg \max_c \ssub{p}{c}$ as $\ssub{\hat{c}}{*}$
      }
    }
    \Return $\arg \max_c \ssub{p}{c}$ as $\ssub{\hat{c}}{*}$\;
    \caption{\textsf{Passive Intent Inference} \label{alg:nd}}}
\end{algorithm}

\subsection{Proactive Solicitation}

\begin{figure}
  \includegraphics[scale=0.75]{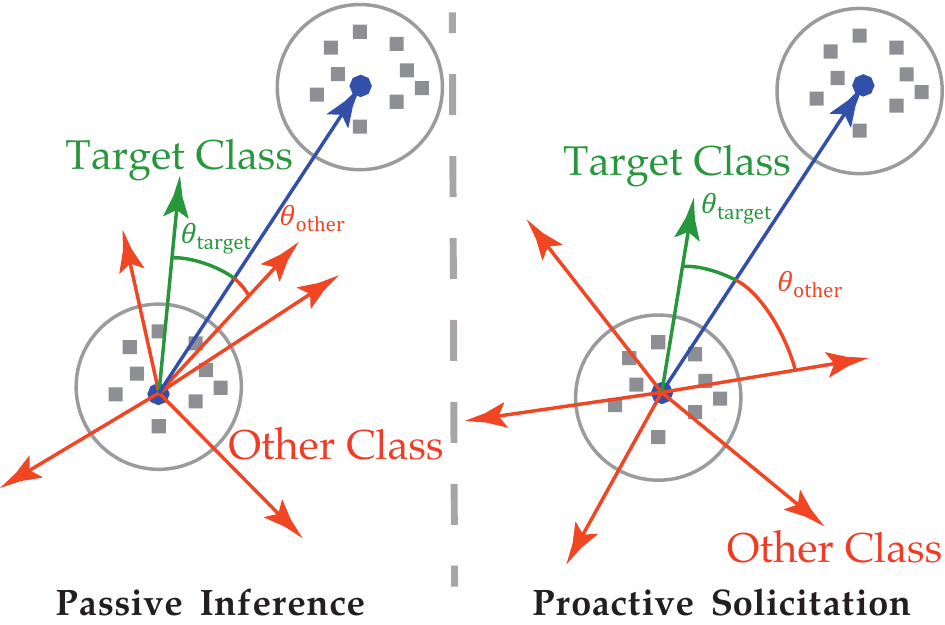}
  \caption{Passive inference versus proactive solicitation}
  \label{fig:active_figure}
\end{figure}

As illustrated in Figure\mref{fig:active_figure}, because the adversary performs descent based on the estimated gradient (Algorithm\mref{alg:blackbox}), the descent direction $\hat{g}(x)$ may deviate from the true gradient direction $g(x)$, which in turn causes possible errors in \system's estimation of $\ay$. The estimate error of $\hat{g}(x)$ mainly results from the bias between $\hat{g}(x)$ and $g(x)$. Take signSGD in \meq{eq:sgd1} as a concrete example; the expectation of  $\hat{g}(x)$ is given by:

\begin{align}
  \label{eq:sgd2}
  \begin{split}
    \sE[\hat{g}(x)]&=\frac{1}{\sigma}\sE_u\left[\left(f\left(x+\sigma u\right)-f\left(x\right)\right)u\right] \\
    &= \sE_u[\left(g(x)\cdot u\right) u] + \sE_u[R(x,u) u] \\
  \end{split}
\end{align}
Where $u\sim\mathcal{N}(0,I)$ and $R(x,u)$ is the second order remainder of $f(x)$ in its Taylor expansion. We have the following proposition (proof in Appendix).
\begin{prop}
  \label{gauss_projection}
  For any vector $v$, we have $\sE_u [\left(v\cdot u\right) u]=v$, where $u\sim\mathcal{N}(0,I)$.
\end{prop}

Applying this proposition to \meq{eq:sgd2}, the first term becomes $g(x)$, which is the true gradient. Yet, due to the high-order term $R(x,u)$, \meq{eq:sgd2} is a biased estimate of $g(x)$. Meanwhile, if we remove the high-order information in the query answers, the adversary would be able to estimate the gradient more accurately, and \system, in turn, would also obtain a more accurate estimation of the adversary's intent. To this end, we define the prediction for a query $x_i$ as:
\begin{align}
  \label{active_eq}
  \hat{f}(x_i)=f(\tilde{x})+ G \cdot (x_i-\tilde{x})
\end{align}
where $\tilde{x}$ is the estimated QOI, $\hat{f}(x_i)$ is the modified prediction on $x_i$, $G$ is a gradient matrix ($G \in \ssup{\sR}{|\gC| \times n}$), in which the $c$-th row is the gradient vector of the class $c$.

We have the following proposition that ensures the similarity of $\hat{f}(x_i)$ and its true value $f(x_i)$ (proof in Appendix):
\begin{prop}
  \label{active_bound}
  If the model $f$ is Lipschitz continuous with $\left\|f(x)-f(y)\right\|\le K\left\|x-y\right\|$, we have
  \begin{align}
    \|\hat{f}(x_i)-f(x_i)\|\le\left(K+\|G\|\right)\left(\sigma \|u\| + B(\tilde{x})\right)
  \end{align}
  where $u \sim \mathcal{N}(0, I)$ and $B(\tilde{x})$ is the bias of the QOI estimate in \msec{sec:estimate}.
\end{prop}

\vspace{3pt}
Applying \meq{active_eq}, the adversary's gradient descent direction now shares the same expectation with the true gradient direction with respect to the target class $\ay$. Yet, due to the estimation variance, the gradient of other classes may be event might be even closer to the adversary's estimate, as shown in Figure\mref{fig:active_figure}. To address this issue, we force the gradient directions of different classes to be orthogonal to each other so that if the adversary follows a descent direction, the target class can be easily distinguished from other classes, as illustrated in Figure\mref{fig:active_figure}.

To this end, the new gradient matrix is defined as follows:
\begin{align}
  \label{active_grad}
  G = (1-\mu) g(\tilde{x}) + \mu M
\end{align}
where the parameter $\mu\in[0,1]$ (dispersion coefficient) controls the magnitude of direction dispersion, and  $M$ is an orthogonal matrix set by $\system$ (an example given below). Specifically, $M$ separates gradients of different classes since each row of $G$ is the gradient of the corresponding class.
\begin{align}
  \label{eq:sample}
  M = \left[
    \begin{matrix}
      1      & 0      & \cdots & 0      & 1      & 0      & \cdots & 0      & \cdots \\
      0      & 1      & \cdots & 0      & 0      & 1      & \cdots & 0      & \cdots \\
      \vdots & \vdots & \ddots & \vdots & \vdots & \vdots & \ddots & \vdots          \\
      0      & 0      & \cdots & 1      & 0      & 0      & \cdots & 1      & \cdots \\
    \end{matrix}
    \right]
\end{align}

Note that \meq{eq:sample} is only one possible instantiation of $M$. Let $c$ be the number of classes and $d$ be the data dimensionality. There are essentially $(c!)^{(d/c)}$ different permutation matrices from which the defender is able to arbitrarily select, creating a prohibitively large number of possibilities for the attacker to explore in order to recover the genuine gradient information.


In conclusion, the proactive solicitation modifies the first-order information (i.e., gradient) while preserving the zeroth-order information. By properly controlling the proportion of gradient information released to the adversary, \system is able to expose the adversary's intent at an early stage while also deterring her from achieving successful attacks. 
\section{Empirical Evaluation}
\label{sec:eval}

Next we empirically evaluate the performance of \system with respect to benchmark datasets, popular DNNs, and state-of-the-art black-box adversarial attacks. The experiments are designed to answer the following questions:
\begin{mitemize}
  \item RQ1: Is \system effective to detect the attacker's intent during an early stage of the attack?
  \item RQ2: Is \system effective against the attacker who purposely attempts to conceal her intent?
  \item RQ3: Is \system effective against a variety of black-box adversarial attacks?
\end{mitemize}
We begin by introducing the experimental setting.

\subsection{Experimental Setting}

\subsubsection*{Datasets} We primarily use four image classification datasets in our evaluation.
\begin{itemize}
  \setlength\itemsep{2pt}
  \item CIFAR10~\cite{cifar} -- it includes 60K 32$\times$32 colored images from 10 classes (e.g., ``ship'');
  \item CIFAR100~\cite{cifar} -- it is essentially CIFAR10 but categorized into 100 fine-grained classes;
  \item ISIC~\cite{Esteva:2017:nature} -- it represents the skin cancer screening task from the ISIC 2018 challenge, in which given $600\times 450$ skin lesion images are categorized into a 7-disease taxonomy (e.g., melanoma);
  \item Mini-VGGface2 -- it is a subset of the VGGface2 dataset~\cite{vggface2}, which consists of $224\times 224$ (center-cropped) color images drawn from 50 individuals (e.g., `Aaron Stanford');.
\end{itemize}

\subsubsection*{DNNs}
We consider 4 pre-trained DNNs as the backend models, VGG13\mcite{vgg} for CIFAR10, VGG16 for CIFAR100, ResNet101\mcite{resnet} for ISIC, and ResNet18 for VGGface2, which respectively attain 92.440\%, 70.470\%, 88.176\%, and 96.175\% accuracy on the testing sets of the corresponding datasets.

\subsubsection*{Attack and Inference Models} We consider 3 state-of-the-art black-box adversarial attack models, NES\mcite{Ilyas:2018:icml}, signSGD\mcite{pmlr-v80-bernstein18a}, and HessAware\mcite{HessAware}, and their adaptive variants (indicated by the superscript of ``A''). More details of the attacks can be found in \msec{sec:background}. We build 3 variants of \system: basic (\system) -- which passively observes the attacker's queries; robust (\amr) -- which employs robust intent estimation; and robust + proactive (\amrp)-- which adopts both robust intent estimation and proactive intent solicitation.

The experiments are conducted on a Linux server with 4 Quodro RTX 6000 GPUs, 2 Intel Xeon processors, and 384G RAM. All the algorithms are implemented in PyTorch. The default parameter setting is summarized in Table\mref{tab:par_setting}. Note that the HessAware attack is only conducted on CIFAR10 and CIFAR100 dataset due to the infeasible cost of computing Hessian matrices for 224$\times$224 images.

\begin{table}[!ht]{\footnotesize
    \setlength\tabcolsep{1.5pt}
    \setlength\extrarowheight{1pt}
    \begin{tabular}{c|l|cccc}
      \multirow{2}{*}{Parameter} & \multirow{2}{*}{Definition}     & \multicolumn{4}{c}{Dataset}                              \\
      \cline{3-6}
                                 &                                 & CIFAR10                     & CIFAR100 & ISIC & VGGface2 \\
      \hline
      $\alpha$                   & adversary's learning rate       & \multicolumn{4}{c}{0.01}                                 \\
      $\varepsilon$              & norm constraint of perturbation & \multicolumn{4}{c}{0.03}                                 \\
      $n_\mathrm{iter}$          & number of attack iterations     & \multicolumn{4}{c}{10}                                   \\
      \hline
      $n_\mathrm{query}$         & queries per iteration           & \multicolumn{4}{c}{100}                                  \\
      $p_\mathrm{fake}$          & proportion of fake queries      & \multicolumn{4}{c}{0.5}                                  \\
      $\sigma$                   & sampling variance               & \multicolumn{4}{c}{0.001}                                \\
      $\tau$                     & HessAware parameter             & \multicolumn{4}{c}{1}                                    \\
      \hline
      $k$                        & QOI estimation iterations       & \multicolumn{4}{c}{1}                                    \\
      $\mu$                      & dispersion coefficient          & 0.1                         & 0.1      & 0.3  & 0.3      \\
      $\kappa$                   & inference confidence            & \multicolumn{4}{c}{0.6}                                  \\
    \end{tabular}
    \caption{Default setting of key parameters. }
    \label{tab:par_setting}}
\end{table}


\begin{figure*}[!ht]
  \centering
  \epsfig{file = 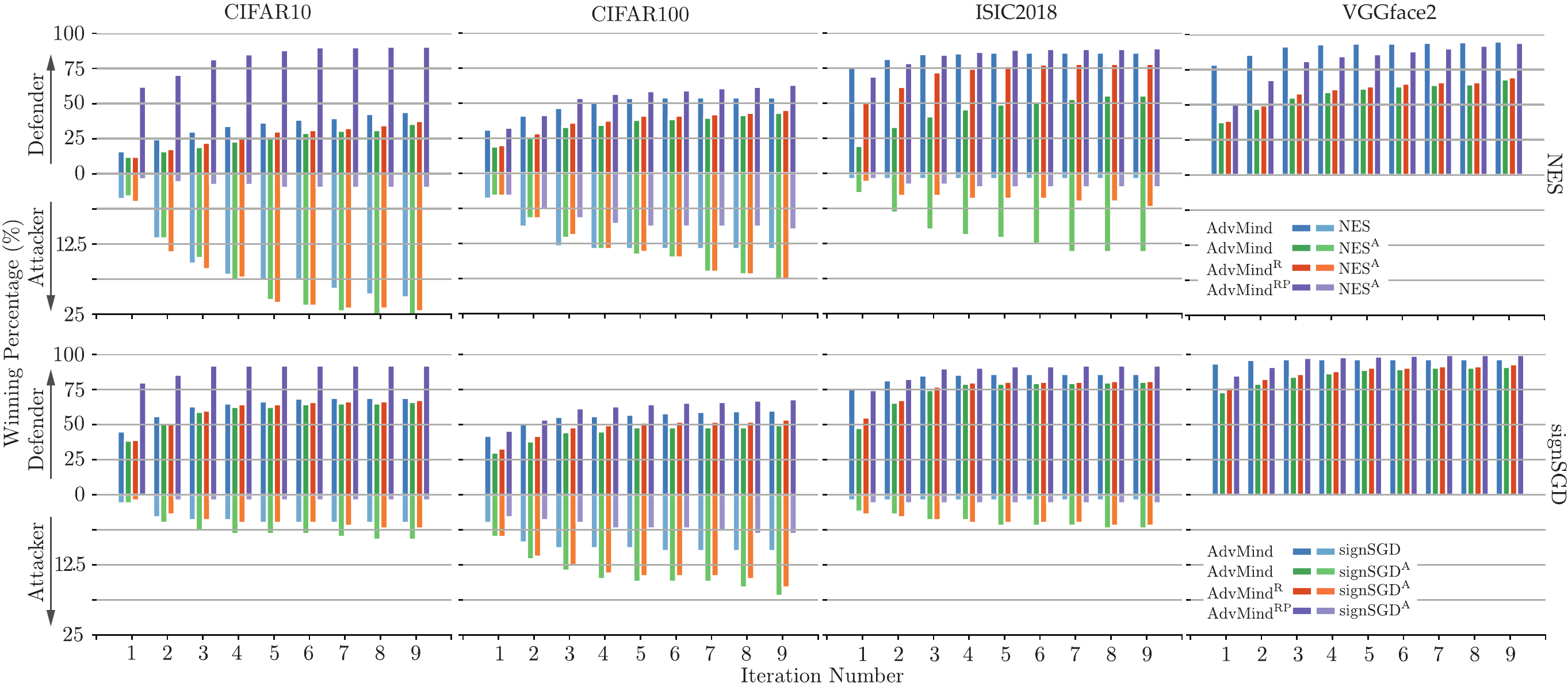, width=180mm}
  \caption{Winning percentages of the attacker and \system in the CTF competition on first order methods with respect to different datasets and DNN models. \label{fig:ctf}}
\end{figure*}

\begin{figure}
  \centering
  \epsfig{file = 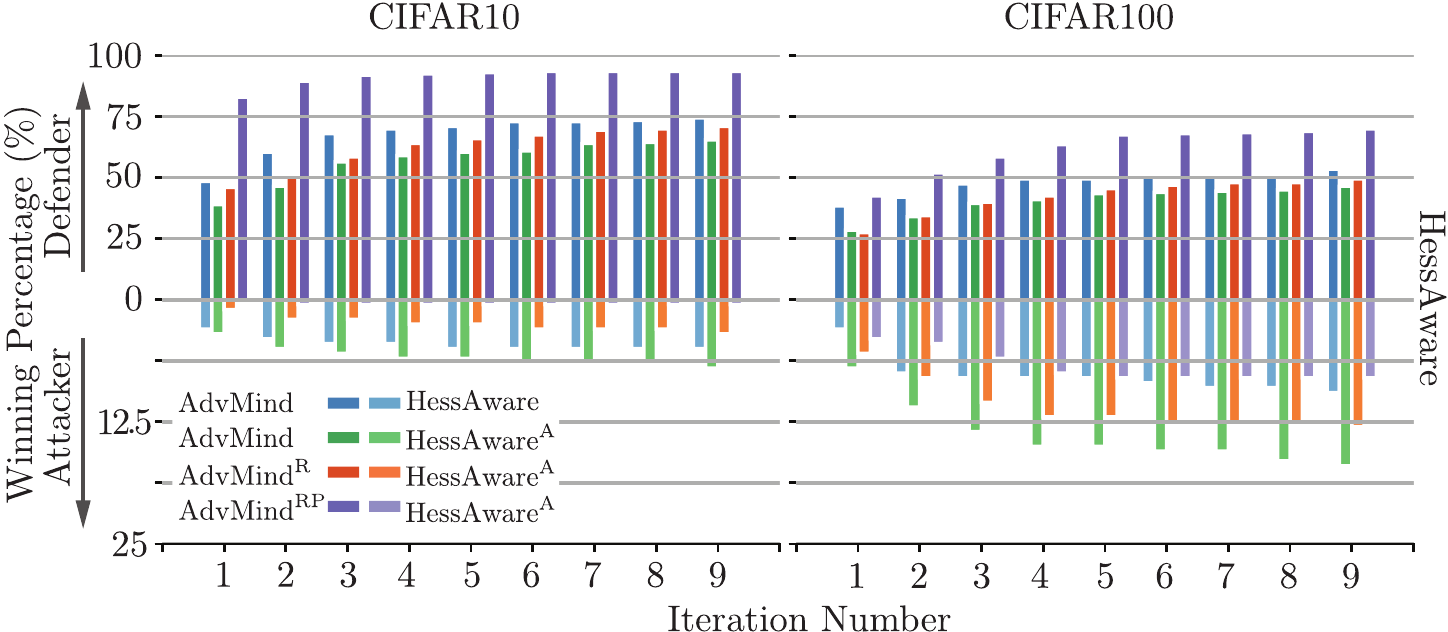, width=90mm}
  \caption{Winning percentages of the attacker and \system in the CTF competition on the second order method with respect to different datasets and DNN models. \label{fig:ctf-hess}}
\end{figure}

\subsection{A CTF Game}
\label{sec:ctf}

To evaluate the performance of \system (the defender) against different black-box attacks and their variants (the attacker). We define the following catch the flag (CTF) game between the attacker and the defender.

\subsubsection*{\bf Setting}  Given the backend DNN $f$, the attacker randomly selects one input $\bx$ from the testing set and a target class $\ay \neq f(\bx)$, and aims to generate an adversarial input $\ax$ that satisfies (i) $\ax \in \mathcal{B}_\varepsilon(\bx)$ and (ii) $f(\ax) = \ay$. At the $k$-th iteration of the attack, the attacker estimates the gradient with respect to the current input $\ssup{x}{(k-1)}$ and then perturbs $\ssup{x}{(k-1)}$ to generate $\ssup{x}{(k)}$; meanwhile, the \system attempts to infer the attacker's target class $\ay$ based on the previous queries. We define that the attacker wins the game if she successfully generates the adversarial input $\ax$ before \system is able to correctly infer $\ay$ with high confidence, and \system wins the game if she correctly infers $\ay$ before the attacker generates $\ax$. Note that if none of them succeeds, it is a ``draw''.

\subsubsection*{\bf Observations}
We measure up to each iteration the percentage of games won by the attacker or the defender, under different configurations for the attacker and the defender. The results on different datasets and DNNs are illustrated in Figure\mref{fig:ctf} and Figure\mref{fig:ctf-hess}, from which we have the following key observations. Note that we aim at early detection so we only focus on the first 10 iterations, during which most attacks haven't succeeded, especially for VGGface2.

\vspace{2pt}
O1: Basic \system is able to infer the attacker's intent in non-adaptive attacks to a large extent. \system defeats the attacker in around 50\% of the games up to the 10-th iteration on CIFAR10 and CIFAR100, and in around 90\% of the games on ISIC and VGGface2. Its better performance on higher-dimensional datasets may be attributed to the larger divergence of gradient directions among different classes in higher-dimensional spaces. Yet, the success rate of \system increases gradually with the number of iterations, indicating that the passive inference requires a large number of observations to make reliable estimation. Furthermore, basic \system is highly sensitive to the fake queries injected by the attacker. Under the adaptive attacks, \system fails to correctly infer the attacker's target before the attacker successfully generates the adversarial examples. The detection rate drops around 10\% on CIFAR10 and CIFAR100 and about 30\% on ISIC and VGGface2. This may be explained by that basic \system relies on na\"{i}ve mean estimation, which tends to significantly deviate from the real value, under the influence of the injected noise. This deviation is especially evident for higher-dimensional data (e.g., ISIC and VGGface2).


\vspace{2pt}
O2: The robust intent estimation of \amr effectively mitigates the influence of fake queries by focusing on the subset of coherent queries to estimate the query of interest. Meanwhile, at each iteration, the attacks attain lower success rates than the case of non-adaptive attacks against basic \system, which is explained by the smaller number of queries useful for gradient estimation.


\begin{figure*}[!ht]
  \centering
  \epsfig{file = 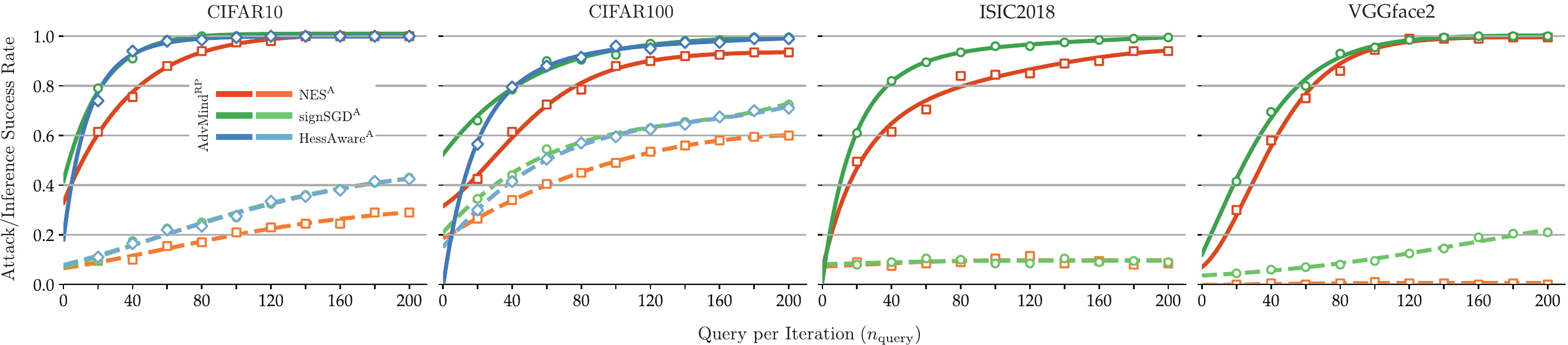, width=180mm}
  \caption{Attack/inference success rate as a function of the number of queries per iteration $n_\mathrm{query}$. \label{fig:query_number} }
\end{figure*}

\begin{figure*}[!ht]
  \centering
  \epsfig{file = 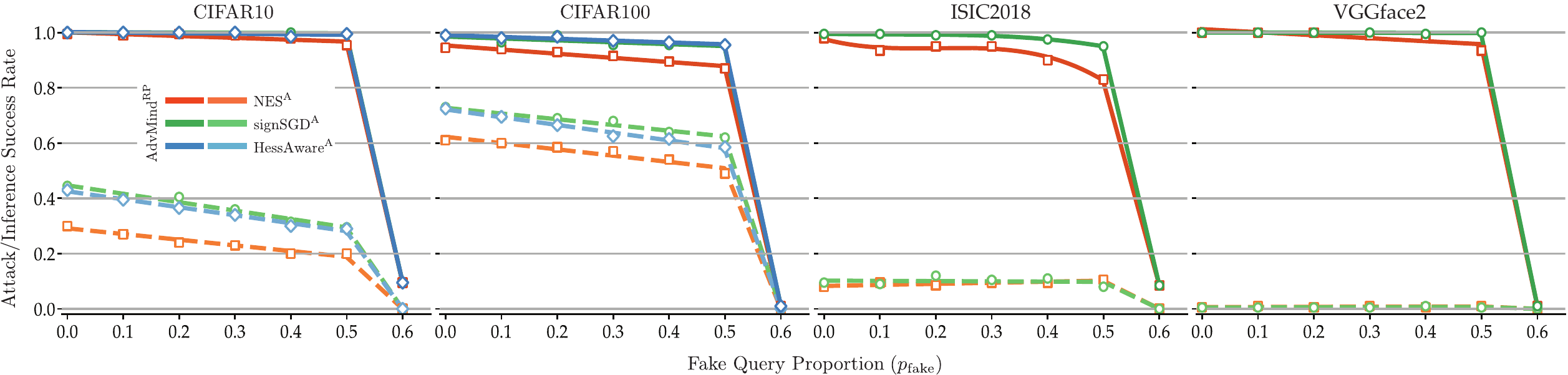, width=180mm}
  \caption{Attack/inference success rate as a function of the proportion of fake queries $p_\mathrm{fake}$. \label{fig:fake_query} }
\end{figure*}


\vspace{2pt}
O3: The proactive intent solicitation of \amrp further significantly improves the inference accuracy, especially during the early stages of the attacks. Across most cases, by the 3rd iteration, \amrp is capable of identifying the attacker's true target with approximately 75\% accuracy even against the adaptive attacks. Meanwhile, because of the dominant performance of \amrp, the attacker's winning percentage is kept close to 0 during the first 10 iterations under all the settings.

\vspace{2pt}
O4: Overall, \system's performance seems agnostic to different datasets, DNNs, or black-box attacks. In comparison, the attacker's effectiveness varies significantly with the concrete setting. For instance, at the 10-th iteration, the winning percentages of most adaptive attacks on ISIC2018 decrease by around 12.5\% compared with their counterparts on CIFAR10. This variations may be explained by that under the black-box setting the adaptive attacks tend to require more iterations to craft adversarial examples in higher-dimensional spaces.

\subsection{Impact of Key Factors}

Next we evaluate the impact of different factors on \system's performance, which shed light on the optimal operation of \system. By default, we use the robust+proactive variant of \system (\amrp) in our evaluation. Unless noted otherwise, all the parameters are set with their default values in Table\mref{tab:par_setting}. To evaluate the inference of \system and the attack effectiveness, we evaluate the inference success rate and the attack success rate up to the 10-th iteration. Note that to factor out the mutual influence, here we evaluate the attacker's and \system's success rates independently.

\subsubsection*{\bf Number of Queries per Iteration $n_\mathrm{query}$} We first measure the impact of the number of queries issued by the attacker per iteration ($n_\mathrm{query}$). Figure\mref{fig:query_number} illustrates the attacker's and \system's success rates on different datasets and DNNs as $n_\mathrm{query}$ varies from 20 to 200. Note that we keep the proportion of fake queries $p_\mathrm{fake}$ constant ($p_\mathrm{fake} = 0.5$) in all the experiments here.

As expected, across all the settings, a large number of queries allow the attacker to generate adversarial inputs more effectively. For instance, on CIFAR10, as $n_\mathrm{query}$ grows from 20 to 200, the attack success rate increases from about 0.1 to 0.4. At the same time, with more queries, \system is also enabled to estimate the attacker's query of interest more accurately, leading to higher inference accuracy. For instance, on CIFAR10, the inference accuracy increases from around 0.7 to 1.0. Therefore, the attacker faces the dilemma between attack effectiveness and intent exposure.



\begin{figure*}[!ht]
  \centering
  \epsfig{file = 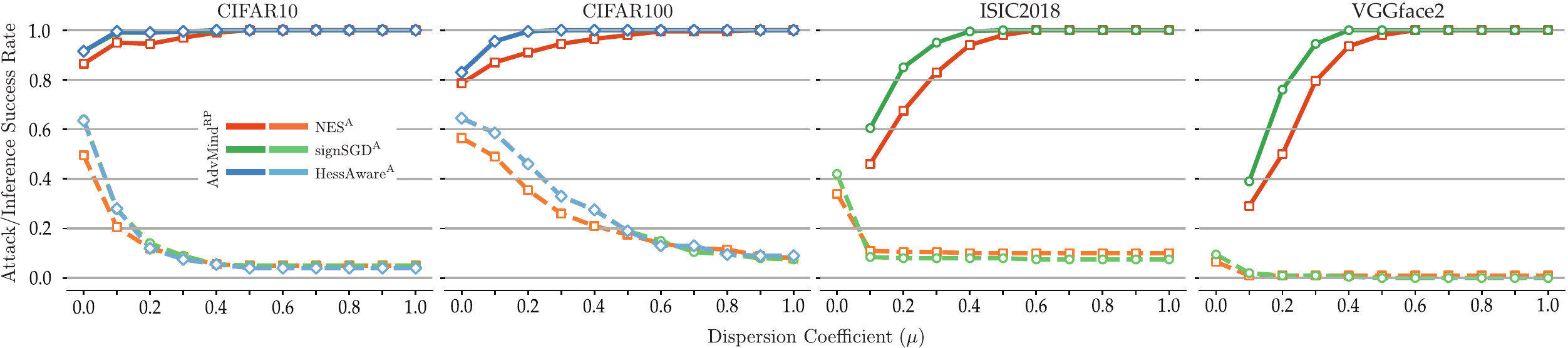, width=175mm}
  \caption{Attack/inference success rate as a function of the dispersion coefficient $\mu$. \label{fig:dispersion} }
\end{figure*}

\subsubsection*{\bf Proportion of Fake Queries $p_\mathrm{fake}$} In this set of experiments, with $n_\mathrm{query} = 100$ fixed, we vary the proportion of fake queries $p_\mathrm{fake}$ to evaluate its impact on the attacker's and \system's performance. Figure\mref{fig:fake_query} plots the results.

It is observed that for both attacker and \system, there exists a critical threshold of $p_\mathrm{fake}$: once $p_\mathrm{fake}$ exceeds this threshold, the attack/inference success rate sharply decreases. For instance, on CIFAR10, if $p_\mathrm{fake} \geq 0.6$, the success rate of \system drops to around 0.1; once $p_\mathrm{fake} \geq 0.6$, the attack success rate drops to around 0. Interestingly, the accuracy drop is larger for \system than that for the attacker. This may be intuitively explained as follow: $p_\mathrm{fake}$ mainly affects the attacker's estimation about the gradient information and \system's estimation about the query of interest; yet, as \system needs first to differentiate fake and true queries (which is set by the attacker), $p_\mathrm{fake}$ tends to have a larger impact on \system. Thus, to hide her intent, the adversary's strategy may be to set $p_\mathrm{fake}$ sufficiently large (e.g., $p_\mathrm{fake} \geq 0.6$), which however significantly increases the attack cost (measured by the number of queries) and reduces the attack success rate.

\subsubsection*{\bf Dispersion Coefficient $\mu$}

For \system, the parameter $\mu$ controls the weight of the dispersion matrix $M$ in \meq{active_grad}. A large $\mu$ implies that the gradient directions with respect to different classes tend to be approximately orthogonal to each other, therefore highly indicative of the adversary's target.

Figure\mref{fig:dispersion} shows the attack and inference success rates as functions of the dispersion coefficient. It is noted that across all the settings, \system's effectiveness increases with $\mu$, while the attack's success rate drops accordingly. This is due to that (i) the orthogonality of the gradient directions increases with $\mu$, making it increasingly easy for \system to identify the adversary's gradient descent direction, and (ii) meanwhile the  error of the gradient information estimated by the adversary grows with $\mu$, making it increasingly difficult to find  adversarial inputs.

However, it is worth pointing out that as $\mu$ increases, the gradient information estimated using the query results becomes less and less informative for the adversary to craft adversarial inputs. Therefore, the adversary may abandon using the query results as $\mu$ reaches a critical threshold. To incentivize the adversary to use the gradient information, the defender may keep $\mu$ small  (e.g., $\mu = 0.1$) in operating \system.



\section{Related Work}
\label{sec:literature}

In this section, we survey the literature relevant to this work.


\vspace{2pt}
{\bf White-Box Attacks and Defenses --} Due to their use in security-critical domains, deep neural network (\dnn) models are increasingly becoming the targets of malicious attacks. Most existing work focus on the white-box setting, in which the adversary has full access to the model information (e.g., its architecture and parameters). One line of work develops new evasion attacks against \dnn models\mcite{goodfellow:fsgm,madry:iclr:2018}. Another line of work attempts to improve \dnn resilience against such attacks by inventing new training and inference strategies\mcite{Ma:iclr:2018}. Yet, such defenses are often circumvented by even powerful attacks\mcite{carlini:sp:2017} or adaptively engineered adversarial inputs\mcite{Athalye:2018:icml}, resulting in a constant arms race between adversaries and defenders\mcite{Ling:2019:sp}.

\vspace{2pt}
{\bf Black-Box Attacks and Defenses --} The increasing popularity of publicly accessible predictive APIs\mcite{model-stealing} has spurred research on black-box adversarial attacks in which the adversary has limited or no knowledge about the target DNN. The first class of attacks leverages the property of transferability\mcite{szegedy:iclr:2014}: certain adversarial inputs crafted against one \dnn model are found effective against another model; thus the adversary is able to generate adversarial inputs based on a surrogate model and apply them on the target model\mcite{liu2016delving}. To defend against such attacks, the method of ensemble adversarial training\mcite{Tramer:2018:iclr} has been proposed recently, which trains a \dnn model using data augmented with adversarial inputs crafted on other models.

Recent work notes that adversarial inputs for surrogate  do not always transfer to the target model, especially when conducting targeted attacks\mcite{chen2017zoo,Narodytska:2017:cvpr}. This line of attacks instead constructs adversarial inputs by estimating the gradient through the target DNNs with coordinate-wise finite difference methods\mcite{Ilyas:2018:icml,liu2018signsgd,HessAware}. The research on defending against query-based black-box attacks is still limited. The recent work\mcite{blackbox-detection} proposes a stateful method to detect query-based black-box attacks by measuring the relationships (e.g., similar but non-identical) of a sequence of queries.


\vspace{2pt}

This work complements the existing work by focusing on identifying the adversary's intent during an early stage of the attack. Meanwhile, early-stage recognition of the adversary's intent represents a long-standing challenge for the security community in general\mcite{inbook,Liu:2005:IMI,Manshaei:2013:GTM}. This work studies this problem within the context of adversarial attacks against DNN models. The findings point to a new direction of mitigating black-box adversarial attacks.

\section{Discussion and Conclusion}
\label{sec:conclusion}

In this paper, we present \system, a new class of models for inferring adversary intent in black-box adversarial attacks. Combining robust intent estimation and proactive intent solicitation, \system is able to reliably identify the adversary's query of interest and accurately detect the adversary's target class in an early stage of the attack, which facilitates to deploy proper mitigation strategies and to perform prompt remediation against such threats. The empirical evaluation with respect to benchmark datasets, popular DNNs, and state-of-the-art attacks validates the efficacy of \system.


Yet, \system is limited in the following aspects, opening up several avenues for further research. 
First, we assume that a single adversary launches the attack and therefore can be easily detected\mcite{blackbox-detection}. In practice, the attack can be performed by multiple adversaries in a coordinated manner. It is critical to extend \system to more complicated settings. Second, we mainly focus on query-based attacks. There are other types of black-box attacks (e.g., surrogate model\mcite{liu2016delving}). As they do not require query access during drafting adversarial inputs, \system is unable to infer the adversary's target in such attacks. Finally, after identifying the adversary's intent, the next step is to perform effective mitigation or remediation. It is beneficial to integrate \system with model enhancement methods (e.g., adversarial training\mcite{madry:iclr:2018}) to provide end-to-end protection.

\newcommand{\bibpre}{bibs}

\bibliographystyle{ACM-Reference-Format}
\bibliography{\bibpre/aml,\bibpre/debugging,\bibpre/general,\bibpre/ting,\bibpre/optimization,main}

\newpage

\section*{Appendix}

\subsection*{A. Symbols and Notations}

\begin{table}[!ht]{\small
    \centering
    \begin{tabular}{c|l}
      Notation            & Definition                              \\
      \hline
      \hline
      $f$                 & DNN model                               \\
      $\ell$              & loss function                           \\
      $\bx$               & original input                          \\
      $\ax$               & adversarial input                       \\
      $\ay$               & adversary's target class                \\
      \hline
      $\alpha$            & adversary's learning rate               \\
      $g(x)$              & gradient of $\ell$ with respect to $x$  \\
      $\hat{g}(x)$        & adversary's estimate of $g(x)$          \\
      $\{\ssup{u}{(k)}\}$ & samples of standard normal distribution \\
      $n_\mathrm{iter}$   & number of attack iterations             \\
      $p_\mathrm{fake}$   & proportion of fake queries              \\
      $\varepsilon$       & norm constraint of perturbation         \\
    \end{tabular}
    \caption{List of symbols and notations. \label{tab:symbol}}}
\end{table}

\subsection*{B: Separability of Consecutive Iterations}
\label{sec:cluster}

\begin{figure}[!ht]
    \centering
    \epsfig{file = 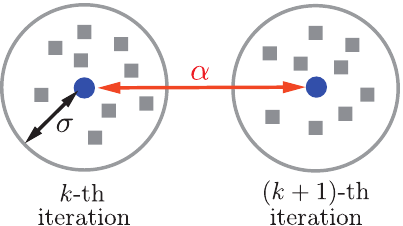, width=50mm}
    \caption{Separability of adversary's queries from the $k$-th and $(k+1)$-th iterations.\label{fig:cluster}}
  \end{figure}

To infer the adversary's intent, \system separates the queries of different iterations. We now justify the feasibility of separating queries from consecutive iterations. In a query-based black-box attack, at each iteration, the adversary samples queries around a QOI to estimate its gradient; the queries thus form a cluster, as illustrated in Figure\mref{fig:cluster}, The sampling variance $\sigma$ determines the cluster radius, while the step length $\alpha$ (corresponding to the learning rate in the adversarial attack) determines the distance between consecutive clusters. Next we prove that in order for the adversary to accurately estimate the gradient, $\alpha$ needs to significantly larger than $\sigma$ (i.e., two consecutive clusters are well separated).  

First, the upper-bound of the finite sample convergence rate of zero-order stochastic optimization can be written as\mcite{wibisono:nips:2012}:
    \begin{align}
        \label{cluster_inequality}
        \begin{aligned}
            \mathbb{E}\left[f(\ssup{\hat{\theta}}{(k)})-f(\theta_*)\right]\leq & 2 \frac{RG\sqrt{s(d)}}{\sqrt{k}}\max \{\alpha, \alpha^{-1}\}        \\
                                                                               & +\alpha u^2 \frac{RG\sqrt{s(d)}}{k} +u\frac{RG\sqrt{s(d)}\log k}{k}
        \end{aligned}
    \end{align}
    where $\theta$ is the optimized parameter, $R$ is the Bregman divergence upper-bound in mirror descent, $G$ is the expectation upper-bound of gradient, $k$ is the number of iterations, $d$ is the data dimensionality, $s(d)$ is a $d$-dependent constraint for the sampling distribution, $u$ is the perturbation magnitude, and $\alpha$ is the step length. Specifically, $s(d)$ is defined as follows. Let $Z$ be sampled according to the distribution $\mu$, where $\mathbb{E}[ZZ^\top]=I$; there exits a constant $s(d)$ such that for any vector $g\in\mathbb{R}^d$, $\mathbb{E}[\|\langle g,Z\rangle Z\|^2_*]\leq s(d)\|g\|^2_*$.

    In our case, $u\sim\mathcal{N}(0,I)$, we have $s(d)=d+3$, which can be derived as follows:
    \begin{align}
        \begin{aligned}
            \mathbb{E}[\|\langle g,u\rangle u\|^2] & =\mathbb{E}\left[\sum_j{\left(\sum_i{g_i u_i}\right)^2 u_j^2}\right]=\mathbb{E}\left[\sum_{ijk}{g_i g_k u_i u_j^2 u_k}\right] \\
                                                   & =\mathbb{E}\left[\sum_i{g_i^2 u_i^4}+\sum_i{g_i^2u_i^2\sum_j{u_j^2}}\right]                                                   \\
                                                   & =\sum_i{g_i^2 3}+\sum_i{g_i^2 \sum_j 1}                                                                                       \\
                                                   & =(d+3)\|g\|^2
        \end{aligned}
    \end{align}

Applying \meq{cluster_inequality} to our setting, we have the following upper-bound:
\begin{align}
    \label{cluster_formula}
      \mathbb{E}\left[f(\ssup{\hat{x}}{(k)})-f(x_*)\right]\leq(\frac{2\sqrt{k}}{\alpha}+\alpha \sigma^2+\sigma \log k) \frac{RG\sqrt{d+3}}{k}
\end{align}
where $\sigma$ is the sample deviation. This upper-bound monotonically increases with $\sigma$ and decreases with $\alpha$. More precisely, it is correlated to the ratio $\frac{\sigma}{\alpha}$. Therefore, to bound the optimization error, the attacker cannot arbitrarily reduce $\alpha$ or increase $\sigma$, which leads to the feasibility of separating consecutive iterations.

For instance, in our evaluation, we set $\sigma=0.001$ and $\alpha=0.01$. The ratio of sum of squares within (SSw) and sum of squares between (SSb) is thus 0.1. To make the separation of consecutive iterations difficult, the attacker may need to adjust $\sigma$ and $\alpha$ to around the same order of magnitude (i.e. $\frac{\sigma}{\alpha} \approx 1$). In this case, the convergence upper-bound is increased by 10 times larger, which significantly impacts the attack effectiveness.

\subsection*{C. Other Proofs}

Next we detail the proofs of the propositions in the paper.

\begin{proof}[\bf Proof of Proposition \ref{gauss_projection}]
    \begin{align*}
        \sE_u\left[\left(v\cdot u\right) u\right]=\sE_u\left[\left(\sum_i {v_i u_i}\right) u\right]
    \end{align*}

    The $j$-th coordinate is
    \begin{align*}
        \begin{split}
            &\sE_u\left[\left(\sum_i {v_i u_i}\right) u_j\right]=\sE_u\left[v_j u_j^2+\sum_{i\neq j} {v_i u_i u_j}\right] \\
            =&v_j \sE_{u_j}\left[u_j^2\right]+\sum_{i\neq j} {v_i \sE_{u_i, u_j}\left[u_i u_j\right]}= v_j\\
        \end{split}
    \end{align*}

    Which implies that $\sE_u\left[\left(v\cdot u\right) u\right]=v$.
\end{proof}

\begin{proof}[\bf Proof of Proposition \ref{active_bound}]

    Given that $\hat{f}(x_i)=f(\tilde{x})+\lambda G\cdot (x_i-\tilde{x})$, apply triangle inequality:
    \begin{align*}
            & \left\|\hat{f}(x_i)-f(x_i)\right\|                                                            \\
        =   & \left\|f(\tilde{x_i})+\lambda G\cdot (x_i-\tilde{x})-f(x_i)\right\|                           \\
        \le & \left\|f(\tilde{x})-f(x_i)\right\|+\left\|\lambda G\cdot(x_i-\tilde{x})\right\|               \\
        \le & \left\|f(\tilde{x})-f(x_i)\right\|+\lambda\left\|G\right\| \cdot \left\|x_i-\tilde{x}\right\| \\
    \end{align*}

    From the Lipschitz assumption: $\left\|f(x)-f(y)\right\|\le K\left\|x-y\right\|$, we can compute the bound as follows:
    \begin{align*}
        \begin{split}
            &\left\|f(\tilde{x})-f(x_i)\right\|+\lambda\left\|G\right\| \cdot \left\|x_i-\tilde{x}\right\| \\
            \le&K\left\|x_i-\tilde{x}\right\|+\lambda\left\|G\right\| \cdot \left\|x_i-\tilde{x}\right\| \\
            =&\left(K+\lambda\left\|G\right\|\right)\left\|x_i-\tilde{x}\right\| \\
        \end{split}
    \end{align*}
    Denote $x$ as the real QOI, we have
    \begin{align*}
        \begin{split}
            &\left(K+\lambda\left\|G\right\|\right)\left\|x_i-\tilde{x}\right\|\\
            \le&\left(K+\lambda\left\|G\right\|\right)\left(\left\|x_i-x\right\|+\left\|\tilde{x}-x\right\|\right) \\
            =&\left(K+\lambda\left\|G\right\|\right)\left(\sigma \|u\|+ B(\tilde{x})\right)\\
        \end{split}
    \end{align*}
\end{proof}

\end{document}